\documentclass{article}
\usepackage{spconf,amsmath,graphicx}

\usepackage[table]{xcolor}
\usepackage{multirow}
\usepackage{booktabs}
\usepackage{amsmath}
\usepackage{mathtools}
\usepackage{amssymb}
\usepackage{enumitem}
\usepackage{tabularx}
\usepackage{adjustbox}
\usepackage{threeparttable}
\usepackage{subfig}
\usepackage{color}
\usepackage{xcolor}
\usepackage{colortbl}
\usepackage{dsfont}

\newcommand{\figref}[1]{Fig.~\ref{#1}}
\newcommand{\tabref}[1]{Table~\ref{#1}}

\newcommand{\eg}{\textit{e.g.}}
\newcommand{\ie}{\textit{i.e.}}
\newcommand*\rot{\rotatebox{90}}

\newcommand{\etal}{\emph{et al}.\@ }

\definecolor{Gray}{gray}{0.9}
\definecolor{mygreen}{rgb}{0.2, 0.7, 0.1}
\usepackage[pagebackref=true,breaklinks=true,colorlinks,citecolor={mygreen},bookmarks=false]{hyperref}

\makeatletter\renewcommand\paragraph{\@startsection{paragraph}{4}{\z@}
  {.5em \@plus1ex \@minus.2ex}{-.5em}{\normalfont\normalsize\bfseries}}\makeatother

\title{TEMPORAL FLOW MASK ATTENTION FOR OPEN-SET LONG-TAILED RECOGNITION OF WILD ANIMALS IN CAMERA-TRAP IMAGES}
\name{Jeongsoo Kim, Sangmin Woo, Byeongjun Park, Changick Kim}
\address{\small{School of Electrical Engineering, KAIST, Daejeon, Republic of Korea}\\
\small{\{jngsoo711, smwoo95, pbj3810, changick\}@kaist.ac.kr}
}
\begin{document}
\ninept
\maketitle
\begin{abstract}
Camera traps, unmanned observation devices, and deep learning-based image recognition systems have greatly reduced human effort in collecting and analyzing wildlife images. However, data collected via above apparatus exhibits 1) long-tailed and 2) open-ended distribution problems. To tackle the open-set long-tailed recognition problem, we propose the Temporal Flow Mask Attention Network that comprises three key building blocks: 1) an optical flow module, 2) an attention residual module, and 3) a meta-embedding classifier. We extract temporal features of sequential frames using the optical flow module and learn informative representation using attention residual blocks. Moreover, we show that applying the meta-embedding technique boosts the performance of the method in open-set long-tailed recognition. We apply this method on a Korean Demilitarized Zone (DMZ) dataset. We conduct extensive experiments, and quantitative and qualitative analyses to prove that our method effectively tackles the open-set long-tailed recognition problem while being robust to unknown classes.
 \let\thefootnote\relax\footnote{This research received support from the National Research Foundation of Korea which is funded by the Ministry of Science and ICT (NRF-2018R1A5A7025409)}
\end{abstract}

\section{Introduction}
\label{sec:intro}
Recently, requirements for large-scale data collection and advanced analytical techniques have led to an increased usage of camera traps~\cite{blount2021covid} --- cameras equipped with motion sensors that can automatically capture wild animals by detecting activity change in their vicinity. Camera traps allow ecological researchers to collect animal images with minimal disturbance to wildlife. At the same time, high-quality image recognition systems actuated by advances in deep learning have enabled researchers to analyze the natural behavior of wildlife at considerably low costs and in short periods~\cite{go2021fine, schneider2019past}. Camera traps are crucial to nest ecology studies~\cite{swann2011evaluating}, identification of threatened species~\cite{johansson2020identification}, and research on habitat and occupation of human-built structures~\cite{rovero2014estimating}.

However, despite tremendous advantages of the camera trap, data collected from the camera traps entail two problems: 1) long-tailed and 2) open-ended distribution of the dataset. By the nature of the ecosystem, animal species in the wild are inherently in a long-tailed distribution. For example, in the food chain, predators are fewer than prey. Moreover, there might be unseen species classes that are not included in the dataset but exist in the real world. For example, the camera trap may not capture animals in hibernation, elusive animals, or endangered species. In these situation, its image recognition system performance may be unfairly underestimated. ~\cite{huang2016learning,kang2019decoupling,zhang2021distribution, liu2021overcome}. A practical recognition system should be able to accurately classify the head (majority) and tail (minority) classes while being robust to the unseen class~\cite{samuel2021generalized,samuel2021distributional,xiang2020learning}.

Existing studies investigating camera trap images tackle the aforementioned data problems by focusing on constructing a sufficiently balanced dataset for each class, applying a bounding box~\cite{schneider2018deep}, and transfer learning~\cite{willi2019identifying}. However, attempts to analyze performance degradation caused by imbalanced data are insufficient~\cite{yu2013automated,norouzzadeh2018automatically,beery2018recognition}. Additionally, due to the open-ended nature of the camera trap images, unknown data can adversely affect classification performance~\cite{beery2018recognition}. Recent studies~\cite{blount2021covid,lu2018attribute,bendale2016towards} show that detecting unknown classes that have never been seen is crucial to monitoring elusive and endangered species. Recent approaches address the open long-tailed recognition problem with the conventional attention mechanism~\cite{vaswani2017attention,liu2019large,yang2019great,zhu2020inflated}. However, they usually perform poorly in camera trap setups where the illumination condition is unstable, the capture space is limited to narrow areas, or the color of the animal is similar to the background.

In this paper, we tackle the open-set long-tailed classification problem using an optical flow-based convolutional attention network, called the Temporal Flow Mask Attention Network (TFMA). TFMA improves feature extraction by utilizing the correlation of sequential 3 frame images that fit the characteristics of the camera trap dataset. The reference image and two optical flow maps provide the location of the moving object, and through selective convolution, the network can attend to a moving object~\cite{liu2020dynamic}. Specifically, we utilize PWC-net~\cite{sun2018pwc} pre-trained on the large synthetic dataset~\cite{DFIB15}, which is an optical flow estimation method that can be learned in real-time end-to-end using CNN. Our dataset is composed of newly collected camera trap images in the Korean Demilitarized Zone (DMZ), namely the DMZ dataset. The dataset is characterized by imbalanced distribution with various noises and it is consists of 12 classes including background and unknown classes. We collected approximately 27,000 sets of 3 consecutive frames by the camera trap image sensor that captures the movement of an object.


We verify the effectiveness of TFMA via comprehensive experiments and ablation studies. The quantitative results show that the proposed network can improve the class imbalance problem and the overall classification accuracy even in an open-set environment. Moreover, we show that applying the meta-embedding technique can boost the performance of the method in open-set long-tailed recognition. Also, qualitative results demonstrate the use of our network can improve class discriminability.

\section{Method}
\label{sec:method}
\begin{figure*}[ht!]
    \centering
    \scriptsize
    \includegraphics[width=1.0\linewidth]{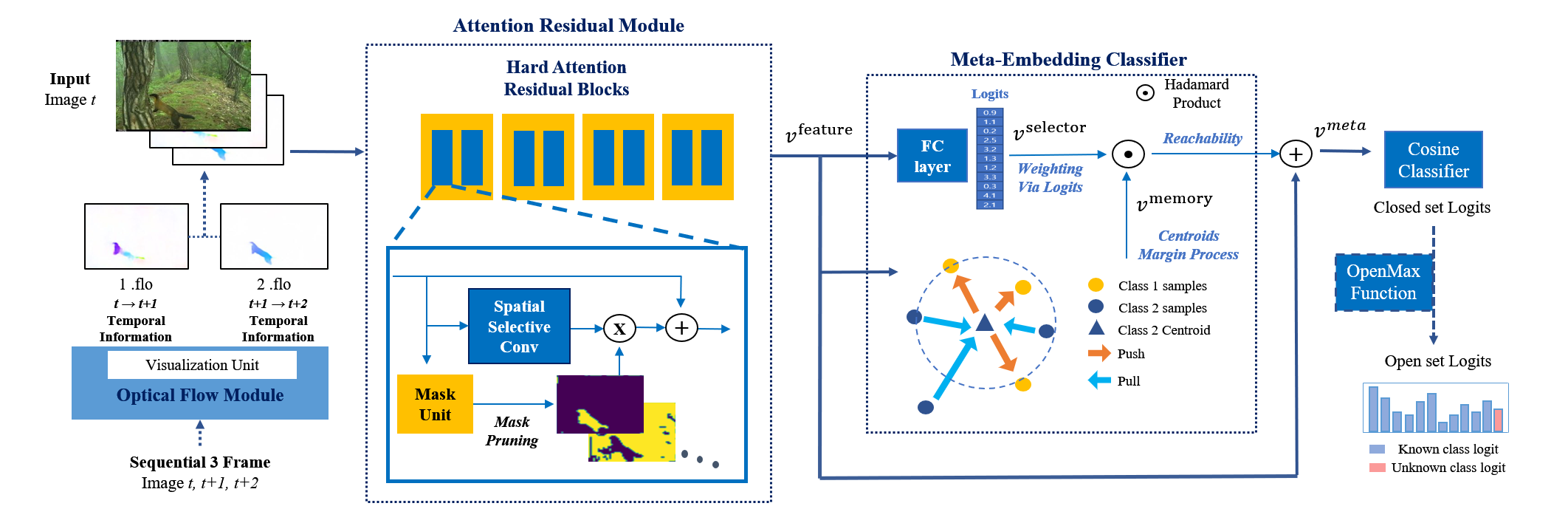}
    \caption{
        Overview of Temporal Flow Mask Attention (TFMA) network.
    }
    \label{fig:overview}
    \vspace{-3mm}
\end{figure*}

An overview of the proposed TFMA network is shown in~\figref{fig:overview}. 

\subsection{Selective Convolution with Mask Operation in Attention Residual Module}

To enhance the selective convolution attention module using input images $t$ with temporal information, we extract the optical flow $({OF}_{1})_{t \rightarrow t+1}$, $({OF}_{2})_{t+1 \rightarrow t+2}$ from consecutive images $t$, $t+1$, $t+2$ using PWC-net~\cite{sun2018pwc}. Before entering the attention residual block, input $x$ is concatenated with temporal flow information. $x = concat(t,{OF}_1, {OF}_2)$. We define the binary step function as a long-tailed estimator used in~\cite{liu2020dynamic, xu2019accurate}, ${S}_{B}(x)$. The derivative of ${S}_{B}(x)$ is defined as follows:

\vspace{-4mm}
\begin{eqnarray}
    \begin{aligned}
        \frac{\delta S_{B}(x)}{\delta x} =\left\{
            \begin{array}{ll}
                2 - 4|x|, &\textrm{-0.4 $\leq$ $x$ $\leq$ 0.4} \\
                0.4, &\textrm{0.4 $\leq$ $|x|$ $\leq$ 1} \\
                0, &\textrm{otherwise} \\
            \end{array}\right.\\
    \end{aligned}
    \label{eq:delta_S_B_x}
\end{eqnarray}

Then, mask operation $M(x)$ used in selective convolution is defined as:

\vspace{-6mm}
\begin{eqnarray}
    \begin{aligned}
        M(x) = S_{B} (\phi(|x|-\theta)-0.5)\,,
    \end{aligned}
    \label{eq:M_x}
\end{eqnarray}
where $\theta$ is the learnable threshold parameter and $\phi$ is the sigmoid function. With the mask operation $M(x)$, input $x$ is fed to the convolutional operation and is added with a skip connection as follows:

\vspace{-4mm}
\begin{eqnarray}
    \begin{aligned}
        output = Conv(x) * M(x) + x\,.
    \end{aligned}
    \label{eq:output}
\end{eqnarray}

\begin{figure}[h!]
    \centering
    \scriptsize
    \includegraphics[width=1.0\linewidth]{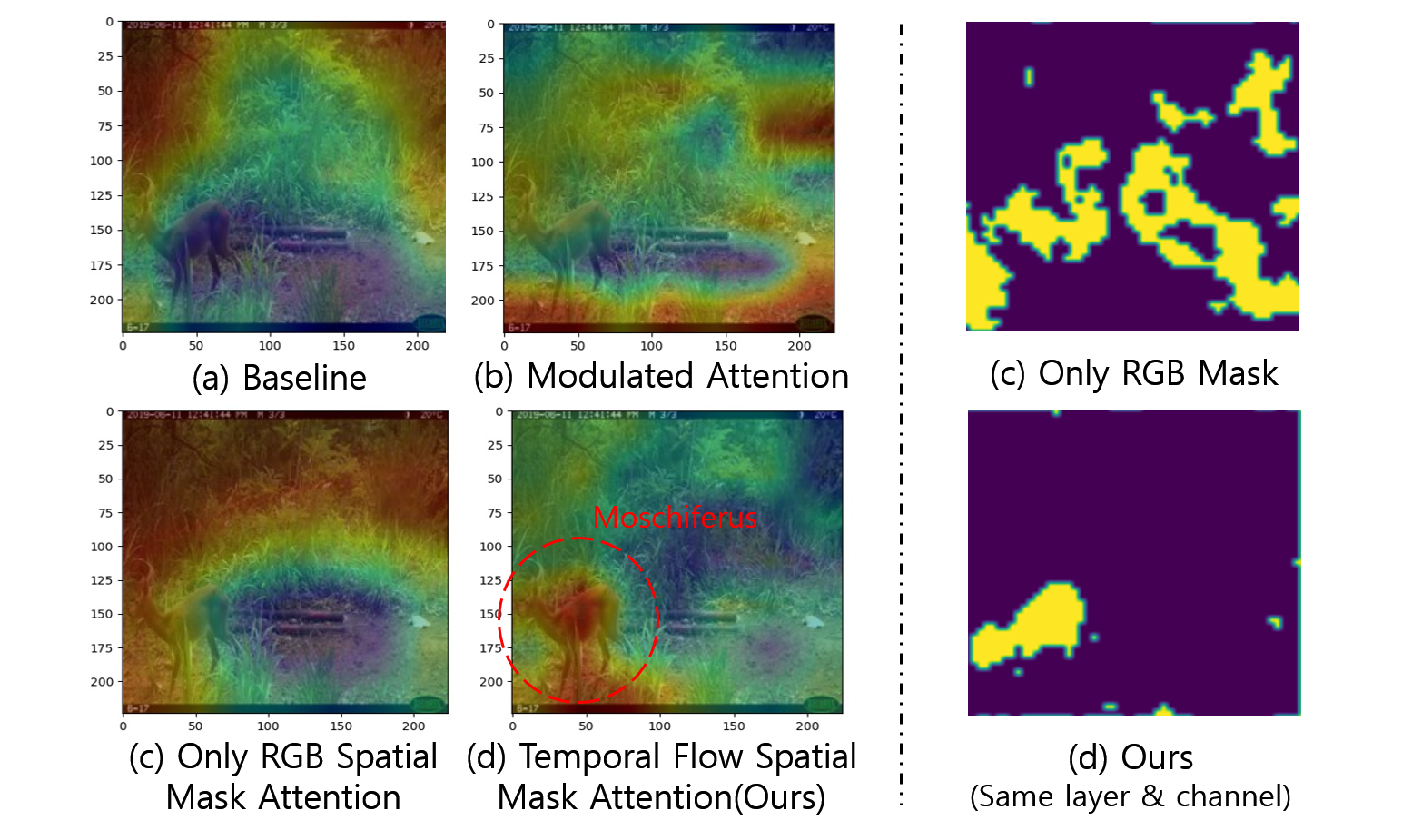}
    \caption{
        Visualization of feature maps and mask attention of animals in the Few class (\eg, endangered species, “moschiferus”).
    }
    \vspace{-2mm}
    \label{fig:visualization}
    \vspace{-2mm}
\end{figure}

With this mask-guided attention module, every residual block is updated with pruned mask weights using the back-propagation procedure. The attention module with optical flow temporal information improves extracting representation in terrifically noisy camera trap images as shown in~\figref{fig:visualization}.

\vspace{-2mm}
\subsection{Meta Embedding Classifier}
After applying the Attention Residual module, we leverage Meta Embedding Classifier~\cite{liu2019large} to fine-tune the performance of TFMA. First, we construct the Concept Selector (${F}_{CS}$) and Hallucinator (${F}_{H}$) module with a simple fully connected layer. Given the output feature vector ${v}^{feature}$ and centroids of the training samples $\{{c}_{j}\}_{j=1}^{N}$, where $N$ is the number of classes, we obtain the memory vector ${v}^{memory}$ and selector vector $v^{selector}$ as follows:

\vspace{-5mm}
\begin{eqnarray}
    \begin{aligned}
        v^{memory} = \sum_{j}^{N} F_{H}(v^{feature})_{j} c_{j}\,,
    \end{aligned}
    \label{eq:v_memory}
\end{eqnarray}

\vspace{-6mm}
\begin{eqnarray}
    \begin{aligned}
        v^{selector} = \textrm{tanh}(F_{CS}({v}^{feature}))\,.
    \end{aligned}
    \label{eq:v_selector}
\end{eqnarray}

Finally, ${v}^{meta}$ combines the output feature and the memory vector with the selector vector. This combination function contributes to fine-tuning the outliers of the samples in head and tail classes.

\vspace{-5mm}
\begin{eqnarray}
    \begin{aligned}
        v^{meta} = \frac{1}{\gamma} \cdot ({v}^{feature} + {v}^{selector} \odot v^{memory})\,,
    \end{aligned}
    \label{eq:v_meta}
\end{eqnarray}
where $\odot$ denotes the Hadamard product, and $\gamma$ denotes the reachability~\cite{savinov2018episodic} to give weight to the distinction between open and tail classes. To train the entire framework, we introduce three loss functions, cross-entropy loss $\mathcal{L}_{CE}$, margin loss $\mathcal{L}_{M}$, and regularization loss $\mathcal{L}_{R}$ defined as follows:

\vspace{-3mm}
\begin{eqnarray}
    \begin{aligned}
        \mathcal{L}_{CE} &(v_{i}^{meta}, y_{i}) = \\
        & y_{i} \textrm{log}(C(v_{i}^{meta})) + (1-y_{i})\textrm{log}(1-C(v_{i}^{meta}))\,,
    \end{aligned}
    \label{eq:L_CE}
\end{eqnarray}
\vspace{-3mm}

\vspace{-3mm}
\begin{equation}
    \resizebox{\linewidth}{!}{
        \begin{minipage}{\linewidth}
            \begin{align}
                &\mathcal{L}_{M} (v_{i}^{meta}, \{{c}_{j}\}_{j=1}^{N}) = \nonumber \\
                &\textrm{ReLU} \left( {\sum_{j}^{N}(
                \mathds{1}_{j = y_i}||v_{i}^{meta}-c_{j}|| -
                \mathds{1}_{j \neq y_i}||v_{i}^{meta}-c_{j}||) + m} \right)\, \nonumber
            \end{align}
        \end{minipage}
    }
    \label{eq:L_LM}
\end{equation}
\vspace{-6mm}

\vspace{-6mm}
\begin{eqnarray}
    \begin{aligned}
        \mathcal{L}_R=\sum_{l}^{L}\sum_{k}^{K} exp(-\theta_{lk}) \,,
    \end{aligned}
    \label{eq:L_R}
\end{eqnarray}
where ${y}$ is the class label, ${L}$ is the number of convolution layers and ${K}$ is the number of channels in the ${l}$-th layer. The classification performance is improved with the use of the cosine classifier $C$~\cite{qi2018low, gidaris2018dynamic} along with the cross-entropy loss $\mathcal{L}_{CE}$. The margin loss $\mathcal{L}_{M}$ is computed between centroids $c_j$ and samples $v_i^{meta}$ and the margin $m$ is set to 10.0. Lastly, we penalize the learnable threshold $\theta$ of the Attention Residual Module with the regularization loss $\mathcal{L}_{R}$. The final loss function is defined as follows:

\vspace{-4mm}
\begin{eqnarray}
    \begin{aligned}
        \mathcal{L}=\sum_{i}^{S}( \mathcal{L}_{CE}^{i}+\lambda_{1} \mathcal{L}_{M}^{i})+\lambda_{2}\mathcal{L}_{R}\,,        
    \end{aligned}
    \label{eq:loss}
\end{eqnarray}
where ${S}$ is the number of mini-batch samples, $\lambda_{1}$ is set to 0.1, and $\lambda_{2}$ is set to 5e-6 via our preliminary experiments on validation set.

\subsection{Open-set Recognition}
Until recently, few OLTR studies investigated open-set and long-tailed problems simultaneously~\cite{liu2019large}. Existing studies only address either the closed-set long-tailed recognition or zero-shot recognition problem. As a result, the evaluation protocols in the literature become inconsistent~\cite{samuel2021generalized, samuel2021distributional, xiang2020learning}. In this study, using the OpenMax function~\cite{scheirer2012toward}, which is typically used in open-set recognition, we investigate whether this challenging circumstance can be solved simultaneously. Specifically, using the OpenMax function, we compare the open-set long-tailed classification performance with the weibull cdf probability using the DMZ dataset that also includes unknown classes. The activation score calculation is modified, but the total activation level remains constant. As the threshold decreases from the optimal value, the uncertainty on tail classes increases. This leads OpenMax to reject more tail classes that are insufficiently learned. Although this improves the confidence to predict the unknown class, it can easily misclassify some tail classes to unknown classes~\cite{scheirer2012toward}. 

By comparing the models with the well-known Receiver Operating Characteristics (ROC), the presented AUROC value indicates the overall accuracy for the unknown class from estimates of all possible thresholds~\cite{kong2021opengan, salehi2021unified}. This demonstrates that our model is highly reliable and capable of minimizing the sensitivity-specificity trade-off. In addition, our model reduces the variance of the intra-class and increases the variance of the inter-class, thereby, ensuring robustness despite degraded performance due to unknown.
 
\section{Experiments}
\label{sec:experiments}

\subsection{Implementation Details}
We perform experiments using the baseline --- ResNet-18~\cite{he2016deep} and the OLTR~\cite{liu2019large}. We employ data augmentation with horizontal flip. The input resolution is resized to 224×224. We train the model for 90 epochs using the SGD optimizer. The weight decay and learning rate are set to 5e-4 and 0.1, respectively.

\vspace{-9pt}
\begin{figure}[h!]
    \centering
    \scriptsize
    \includegraphics[width=1.0\linewidth]{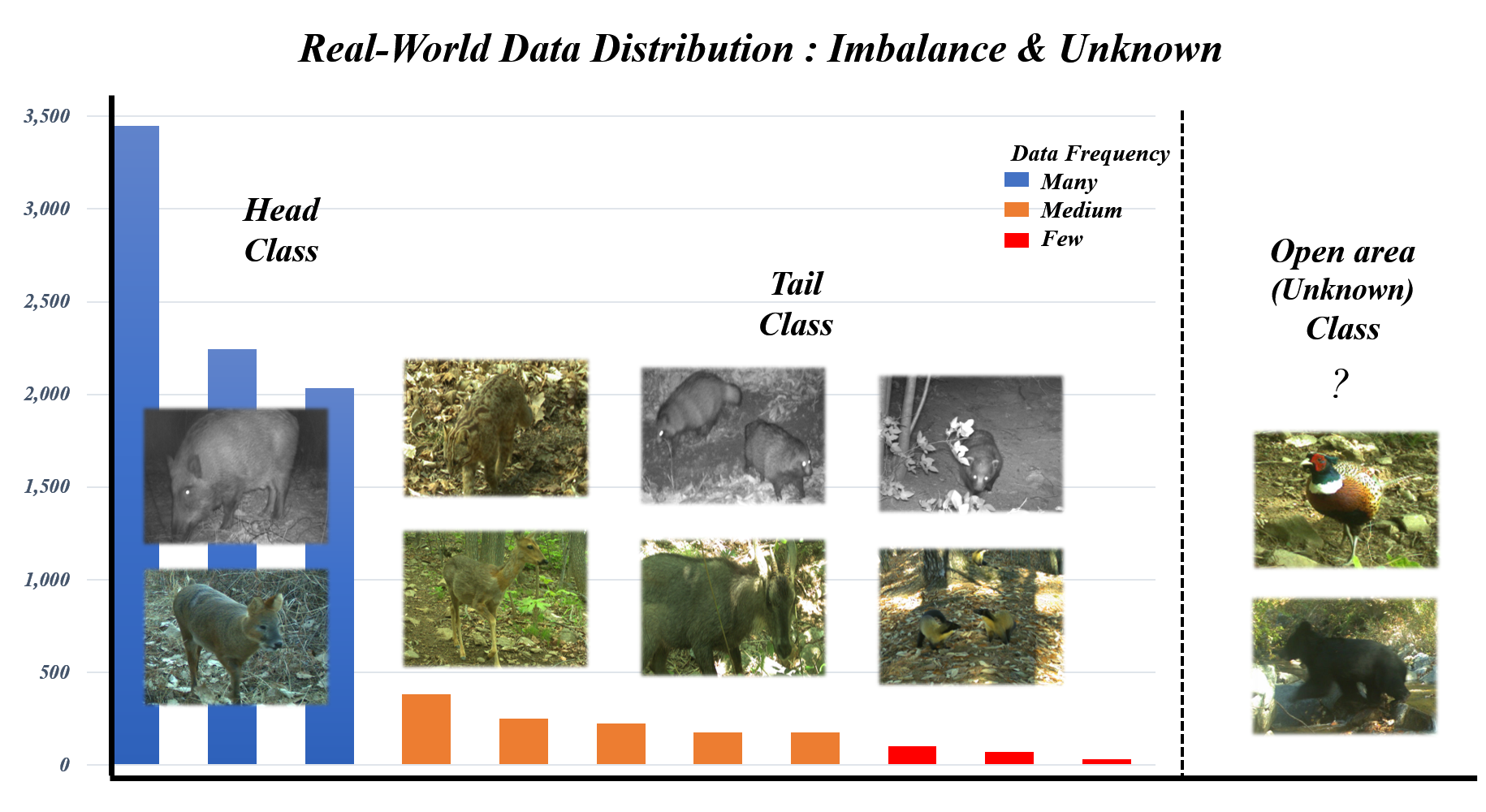}
    \vspace{-11pt}
    \caption{
        Real-world open-set data distribution in the DMZ.    }
    \label{fig:data_distribution}
    \vspace{-5mm}
\end{figure}
\subsection{Open-set Long-tailed Dataset in DMZ}
We collected raw images using the Reconyx HC600 camera trap in the DMZ in collaboration with ecologists from the National Institute of Ecology and scientists from the Anthropocene Research Center of the Korea Advanced Institute of Science and Technology. Unlike Park~\etal\cite{park2022balancing} who focused on the imbalances between domains by splitting the data by domains (\eg, RGB and IR), we expand and reconfigure the dataset to consider a more general open-ended situation. We configured a mammalian dataset including 4 endangered wild animal species (\ie, caudatus, moschiferus, flavigula, bengalensis) belonging to the tail class in the DMZ. We classify 11 classes including a background class as a closed set and 14 classes including 3 Unknown classes as an open-set. Training is carried out only in the closed-set in which the training and testing datasets are split in a 7:3 ratio. In addition, using the day and night background classes of 10,000 photos, the network learns the background noise and camera shake without objects; therefore, it is robust to domain changes and classifies the background into the Many class. Also, similar to~\cite{park2022balancing}, and because most of the image sequences were captured within the identifiable range of the object motion (based on 3 photos), we considered the first 3 frames for each image sequence. Consequently, approximately 27,000 images of 3 sequential frames were extracted. The data sequence frequency distribution is shown in~\figref{fig:data_distribution}. The open-set long-tailed distribution of our DMZ dataset is as follows: \textit{Many Class} (400 or more), \textit{Medium Class} (100 to less than 400) and \textit{Few} ${\&}$ \textit{Unknown Class} (less than 100).


\subsection{Ablation Studies}
We verify the performance of the proposed models under the data configuration using real-world long-tailed circumstances.

\vspace{-3pt}
\begin{table}[h!]
    \centering
    \footnotesize
    \begin{tabular}{ccc|c|ccc}
    \toprule[0.15em]
    \multicolumn{3}{c|}{Imbalanced Closed Test set} & Top-1 Acc. & Many & Medium & Few \\ \midrule \midrule
    \multicolumn{3}{c|}{ResNet-18~\cite{he2016deep} (only RGB)} & 91.41 & 94.05 & 74.97 & 36.61 \\ \midrule
    \multicolumn{3}{c|}{OLTR~\cite{liu2019large} (only RGB)} & 91.81 & 94.34 & 75.10 & 41.53 \\ \midrule
    \multicolumn{2}{c|}{\multirow{2}{*}{\rot{Ours}}} & \cellcolor{Gray}TFMA & 92.13 & 93.17 & 76.68 & 48.09 \\
    \multicolumn{2}{c|}{} & \cellcolor{Gray}TFMA+meta-emb & \textbf{92.85} & \textbf{94.53} & \textbf{77.21} & \textbf{53.01} \\
    \bottomrule[0.15em]
    \end{tabular}
    \caption{Comparison of model accuracies in an imbalanced dataset with different settings (All vs. Many vs. Medium vs. Few classes).}
    \label{tab:accuracy}
\end{table}
\vspace{-3pt}

\vspace{6pt}
\noindent{\textbf{Ablation studies on the closed test set: Long-tailed recognition.}}
The performance comparison results on the imbalanced closed test set are shown in~\tabref{tab:accuracy}. From the table, the classification performance of the ResNet-18~\cite{he2016deep} is highly biased towards the Many class, thereby, adversely affecting the classification performance for the Few class. The OLTR~\cite{liu2019large} tries to solve the long-tailed problem by adding a modulated attention function and a meta-embedding classifier function to ResNet-18, and, this technique tackles the data imbalance problem of the DMZ dataset more effectively. In the camera trap setup, however, the modulated attention module often fails to capture the animal owing to extreme image noises (which can be observed in the feature map of~\figref{fig:visualization} (b)). This also suggests that the DMZ dataset includes data imbalance and problems with the trap images~\cite{beery2021iwildcam}, such as camouflage, blur, close-up shot, fine-grained species, occlusion, and parts only. In addition, when comparing the feature maps \& mask in~\figref{fig:visualization} (c) and (d), the method of applying spatial attention by thresholding only RGB images in the convolution layer focuses on background noise. To solve this problem, the TFMA network utilizes the temporal feature extracted from the optical flow of the object's movement over time to increase attention to the moving object, using the attention module.

The classification accuracy decreases by 1.17\% in the Many class of our TFMA model compared to the OLTR model, whereas it increases by 1.58\% and 6.56\% in the Medium and Few classes, respectively. 
However, although the classification accuracy decreases a little in the Many class of our TFMA model compared to the OLTR model, its accuracy for the tail classes increased significantly. In addition, the classification accuracy of all Many classes can be improved when the embeddings are fine-tuned using TFMA. Therefore, when data are insufficient, it becomes robust against bias due to the background noise by blending temporal information. 



\vspace{6pt}
\noindent{\textbf{Ablation studies on the open test set: Unknown vs Known.}}

\begin{table}[h!]
    \centering
    \footnotesize
    \begin{tabular}{ccc|cc}
    \toprule[0.15em]
    \multicolumn{3}{c|}{Open-set test} & \multicolumn{2}{c}{AUROC} \\ \cmidrule(l){4-5}
    \multicolumn{3}{c|}{(Unknown vs. Known)} & Balanced Test set & Imbalanced Test set \\ \midrule \midrule
    \multicolumn{3}{c|}{ResNet-18 (only RGB)} & 0.4536 & 0.5934 \\ \midrule
    \multicolumn{3}{c|}{OLTR (only RGB)} & 0.6792 & 0.7766 \\ \midrule 
    \multicolumn{2}{c|}{\multirow{2}{*}{\rot{Ours}}} & \cellcolor{Gray}TFMA & 0.6788 & 0.8132 \\
    \multicolumn{2}{c|}{} & \cellcolor{Gray}TFMA+meta-emb & \textbf{0.7744} & \textbf{0.8845} \\
    \bottomrule[0.15em]
    \end{tabular}
    \caption{Comparison of AUROC in open-set recognition.}
    \label{tab:auroc}
\end{table}
\vspace{-10pt}

\vspace{-1mm}
\begin{figure}[h!]
    \centering
    \scriptsize
    \includegraphics[width=1.0\linewidth]{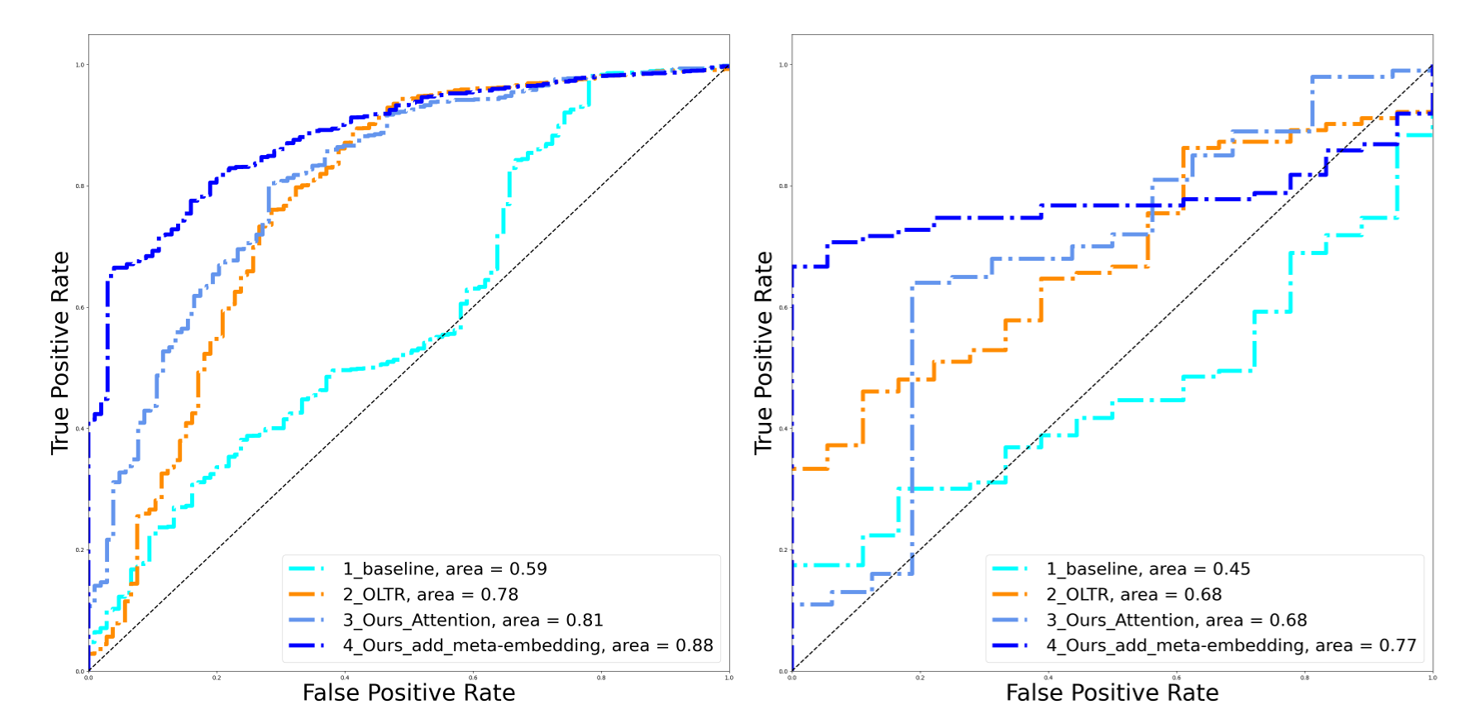}
    \caption{
        ROC curve of open-set recognition with an imbalanced (left) and a balanced (right) test set. 
    }
    \label{fig:roc_curve}
\end{figure}

Since the accuracy of classifying an unknown class in open-set recognition varies depending on the threshold value of the extreme value distribution estimate of the classifier, we verify the reliability of the model for open-set recognition through the ROC curve. As the True Positive Rate (TPR) increases significantly compared to the False Positive Rate (FPR) in the ROC curve, the reliability of the model increases despite the trade-off between sensitivity and specificity. In other words, a classifier is known as a good classifier when the curvature of the graph is closer to the upper left~\cite{bendale2016towards}.

From~\tabref{tab:auroc}, the overall performance is improved in the order of the models proposed by AUROC, and the OLTR model classifies unknown classes better than ResNet-18.
However, despite using the embedding function to reduce outlier bias, TFMA can still improve the overall TPR without the embedding function. This is consistent with existing papers that exclude classifier performance from long-tailed recognition and still realize high-performance improvement even with enhanced representation learning~\cite{samuel2021distributional}. In addition, TFMA fine-tuned with the meta-embedding technique achieves the same experimental results as previous papers on the importance of classifier performance~\cite{kang2019decoupling}. Suffice to say, appropriate fine-tuning, the representation learning step, and the decoupling of the classifier improve the effectiveness of a deep learning network in open-set long-tailed classification. As shown in~\figref{fig:roc_curve} (left), the ROC curve supports the aforementioned results. Our model is robust to the threshold of the OpenMax in an open-ended long-tailed distribution. Figure \ref{fig:roc_curve} (right) shows the results when each class in the balanced testing dataset is randomly composed of 18 images, which is the maximum number of samples in the Few class. The dataset is a few-balanced sample, which makes data sampling highly biased. In this situation, the distribution of the test samples becomes less representative, and the classifier becomes more sensitive to hyperparameters; therefore, the greater the distance from the optimal threshold, the more unreliable the performance comparisons between the networks. 

Nevertheless, the bottom-left part of the curves in~\figref{fig:roc_curve} shows that our model, with the embedding classifier, fine-tuned properly, tends to maintain a stable TPR under rigorous conditions requiring a low FPR owing to an extremely low threshold. Therefore, we confirm that TFMA+meta-emb is more reliable than its counterparts in terms of handling the trade-off between sensitivity and specificity, and can generalize well that ensures a relevant degree of TPR (\ie, sensitivity) performance.



\vspace{6pt}
\noindent{\textbf{Effect of Meta-embedding classifier.}}
\begin{figure}[h!]
\vspace{-3mm}
    \centering
    \scriptsize
    \includegraphics[width=1.0\linewidth]{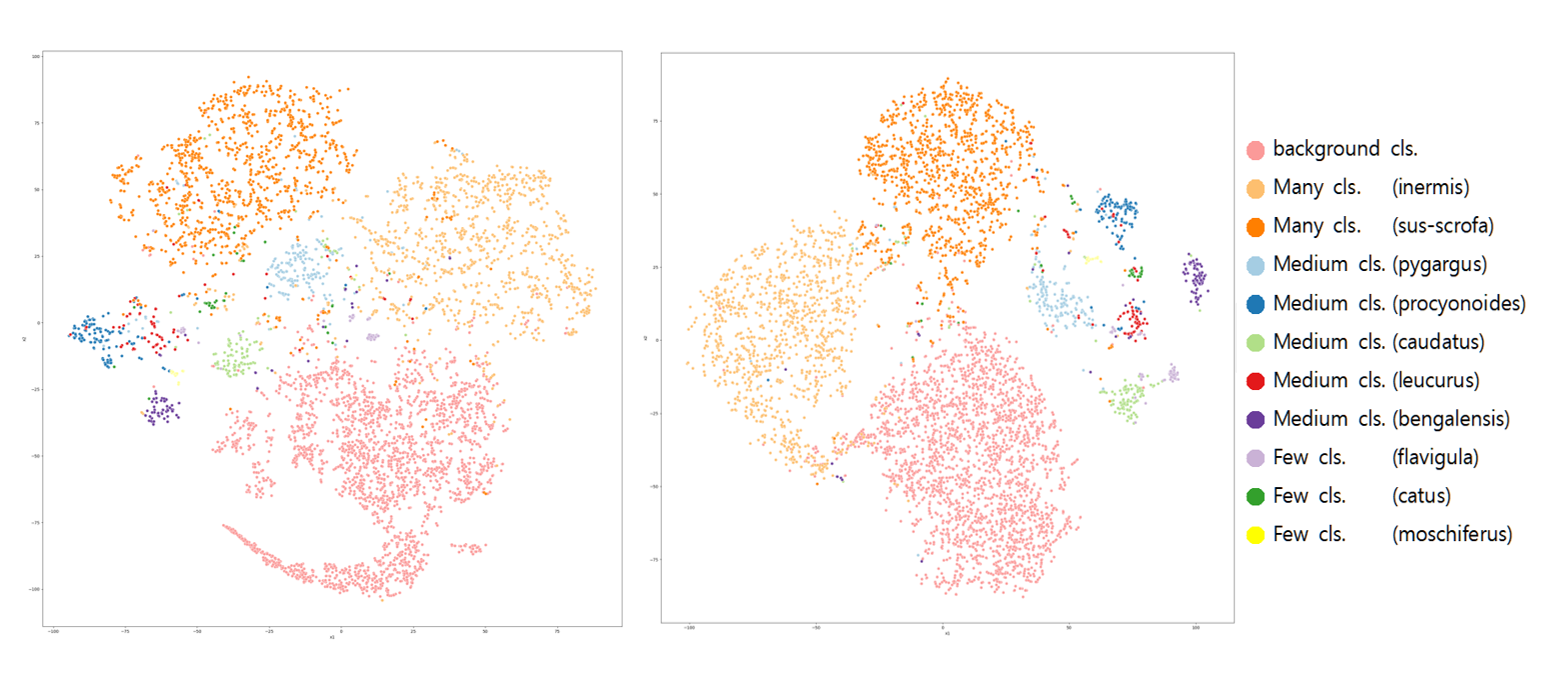}
    \vspace{-3mm}
    \caption{
        t-sne visualization: w/o (left). vs. w/ Meta-embedding.(right)
    }
    \label{fig:tsne}
   
\end{figure}

In~\figref{fig:tsne}, we show the effect of the embedding function by visualizing the output embeddings via t-sne~\cite{van2008visualizing}. With meta-embedding, TFMA learns a reasonable degree of similarity between each sample and class. 
We see that the inter-cluster distinction is clearer and the intra-cluster is compacter in (right) compared to (left), implying that TFMA fine-tuned with meta-embedding generates a more distinctive representation.
By updating the parameters with knowledge distillation~\cite{xiang2020learning,hinton2015distilling}, the meta-embedding classifier learns high-quality representation, thus can generalize well even in the class imbalanced and open-ended distributions.

\section{Conclusion}
\label{sec:conclusion}
To tackle the open-set long-tailed recognition problem, we propose a Temporal Flow Mask Attention (TFMA) network composed of three key components: an optical flow module, an attention residual module, and a meta-embedding classifier. TFMA utilizes temporal information extracted from consecutive frames to adaptively learn attentive representation and predicts the final outputs via a meta-embedding classifier. We applied TFMA to a dataset collected from the DMZ region that consisted of two problems: 1) long-tailed and 2) open-ended distributions. To verify the reliability of TFMA on an open long-tailed setup, we conducted extensive experiments and analyses. The experimental results demonstrate that our model not only improves recognition performance on the tail class but is also robust to the unknown class.



\vfill\pagebreak
\bibliographystyle{IEEEbib}
\bibliography{refs}

\end{document}